\documentclass[letterpaper, 10 pt, conference]{IEEEtran}
\IEEEoverridecommandlockouts
\usepackage{cite}
\usepackage{amsmath,amssymb,amsfonts}
\usepackage{algorithmic}
\usepackage{graphicx}
\usepackage{textcomp}
\usepackage{xcolor}
\usepackage[utf8]{inputenc}
\usepackage[english]{babel}

\usepackage{amsthm}
\usepackage{ wasysym } 
\usepackage[autostyle]{csquotes}
\usepackage{algorithm}
\usepackage{todonotes}
\let\labelindent\relax
\usepackage{enumitem}

\usepackage[font=footnotesize,skip=6pt]{caption}
\usepackage{etoolbox}
\apptocmd{\thebibliography}{\scriptsize}{}{}

\newtheorem{theorem}{Proposition}
\newtheorem*{proposition-1*}{Proposition 1}
\newtheorem*{proposition-2*}{Proposition 2}
\newtheorem*{proposition-3*}{Proposition 3}
\newtheorem{definition}{Definition}
\newtheorem*{problem*}{Problem Statement}

\newtheorem{scenario}{Scenario}
\newtheorem*{scenario*}{Scenario}

\usepackage[colorlinks,citecolor=red,urlcolor=black,bookmarks=false,hypertexnames=true]{hyperref} 

\def\BibTeX{{\rm B\kern-.05em{\sc i\kern-.025em b}\kern-.08em
    T\kern-.1667em\lower.7ex\hbox{E}\kern-.125emX}}
\begin{document}

\title{\LARGE \bf \textcolor{blue}{Extended Version of }Reactive Task Allocation and Planning for Quadrupedal and Wheeled Robot Teaming 
}

\author{
Ziyi Zhou$^{1}$, Dong Jae Lee$^{1}$, Yuki Yoshinaga$^{1}$, Stephen Balakirsky$^{2}$, Dejun Guo$^{3}$, and Ye Zhao$^{1}$
\thanks{$^1$The authors are with the Laboratory for Intelligent Decision and Autonomous Robots, Woodruff School of Mechanical Engineering, Georgia Institute of Technology. {\tt\footnotesize \{zhouziyi, dlee640, yyoshinaga3, yzhao301\}@gatech.edu}.}
\thanks{$^2$The author is with Georgia Tech Research Institute, Atlanta, GA
30318. {\tt\footnotesize \{Stephen.Balakirsky@gtri.gatech.edu}\}.}
\thanks{$^3$The author is with the UBTECH North America Research and Development Center Corporation. {\tt\footnotesize \{dejun.guo@ubtrobot.com}\}.
}
\thanks{This work was supported by UBTECH, NSF grant \#IIS-1924978, \#CMMI-2144309, and Georgia Tech Research Institute IRAD Grant.}
}

\maketitle

\begin{abstract}
This paper takes the first step towards a reactive, hierarchical multi-robot task allocation and planning framework given a global Linear Temporal Logic specification. \textcolor{black}{The capabilities of both quadrupedal and wheeled robots are leveraged} via a heterogeneous team to accomplish a variety of navigation and delivery tasks. \textcolor{black}{However, when deployed in the real world, all robots can be} susceptible to different types of disturbances, including but not limited to locomotion failures, human interventions, and obstructions from the environment. To address these disturbances, we propose task-level local and global reallocation strategies to efficiently generate updated action-state sequences online while guaranteeing the completion of the original task. These task reallocation approaches eliminate reconstructing the entire plan or resynthesizing a new task. \textcolor{black}{To integrate the task planner with low-level inputs}, a Behavior Tree execution layer monitors different types of disturbances and employs the reallocation methods to make corresponding recovery strategies. To evaluate this planning framework, dynamic simulations are conducted in a realistic hospital environment with a heterogeneous robot team consisting of quadrupeds and wheeled robots for delivery tasks.
\end{abstract}


\section{Introduction}

Mobile robots have been extensively investigated and deployed in various service applications such as assembly \cite{halperin2000general}, surveillance, \cite{ulusoy2013optimality} and search and rescue \cite{jennings1997cooperative}. In recent years, quadrupedal robots have been popularized for their superior traversability over \textcolor{black}{unstructured terrains} \cite{biswal2020development}. Nevertheless, even with exceptional locomotion capabilities, legged systems are often unstable, fragile, and less suitable for performing prolonged tasks compared to wheeled robots. However, distinct types of robots can form a heterogeneous team to compensate for their individual disadvantages.

Recent works on multi-robot systems have been focusing on mission planning problems with the assistance of formal languages such as Linear Temporal Logic (LTL) \cite{shoukry2017linear}. Originally proposed for model checking \cite{pnueli1977temporal}, LTL is a powerful tool used in the robotics community with a preponderance of research primarily conducted on wheeled robots \cite{kress2009temporal, decastro2018collision} and legged robots \cite{warnke2020towards, zhao2022reactive, kulgod2020temporal} for task and motion planning. There have also been works  \cite{kress2009temporal,caoleveraging,guo2015multi,tumova2016multi,kantaros2020stylus} on multi-agent systems. However, objectives are explicitly assigned to individual robots rather than having one global specification. This can be challenging for a large team of robots\cite{banks2020multi,kolling2015human}, where in most cases, a global task is simpler to define. Therefore, a simultaneous task allocation and planning (STAP) problem given a global LTL specification attracts more attention.

\begin{figure}[t]
    \centering
    \includegraphics[width=\linewidth]{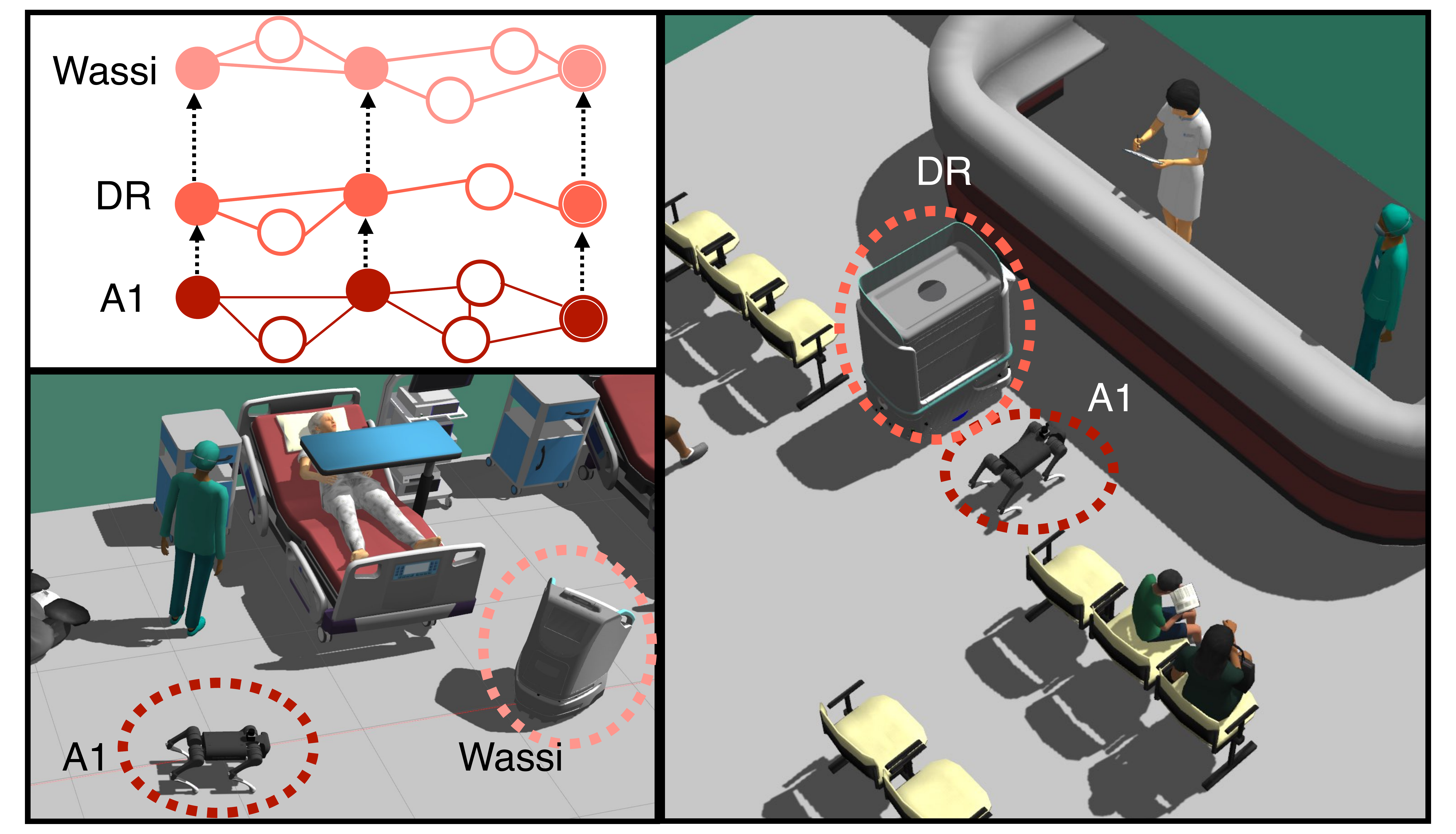}
    \caption{A conceptual illustration of a three robots (Unitree A1, UBTECH DR, UBTECH Wassi) performing tasks in a simulated hospital setting. An illustration of team automaton is shown in the \emph{top left}. The role of this team automaton is described in detail in Fig.~\ref{fig:system_diagram}.}
    \label{fig:vision}
    \vspace{-0.2in}
\end{figure}

A line of research exists where STAP problems have been solved with global LTL specifications. In \cite{chen2011formal,ulusoy2013optimality} a product model is constructed with an exponential complexity, while \cite{schillinger2018decomposition} proposed a team model automaton with a linear complexity, assuming that each robot conducts its task independently. Other works also focused on advanced search algorithms \cite{banks2020multi,schillinger2017multi}, concrete time constraints \cite{leahy2020fast}, and collaborative tasks \cite{luo2021temporal}. However, failure recovery during real-world deployment is rarely studied, \textcolor{black}{especially considering unstructured environments such as rough terrain}. Single and multi-robot scenarios have been demonstrated with disturbances caused by a change in the environment \cite{guo2013revising,guo2013reconfiguration,livingston2012backtracking,li2021reactive} or a failure to perform an action \cite{tumova2014maximally}, but all have been limited to local LTL specifications. The work of \cite{faruq2018simultaneous} is conceptually similar to our goal, where each robot reallocates its tasks during execution failures, based on a Team Markov Decision Process. However, this work manually designs a set of LTL specifications in advance without identifying decomposable parts of a global task specification for optimal task sequences. 
\textcolor{black}{Therefore, task reallocation capabilities during the online execution is desirable for STAP problems given global LTL specifications, which is addressed in this study.} 

In this paper, multiple reallocation approaches to a rich class of online disturbances are proposed to replan on the updated team model given one global LTL specification. 
\textcolor{black}{A local reallocation efficiently outputs a new action sequence only for the problematic robot, while a global reallocation searches on the entire team model and generates optimal action sequences for all robots.} 
Meanwhile, the connection between the high-level reallocation and the low-level disturbance detection is interfaced by a mid-level Behavior Tree (BT). BT is a commonly used graphical and mathematical controller for a robot that enables fault-tolerant task executions \cite{schillinger2018simultaneous,colledanchise2019towards,abiyev2016robot,kim2018architecture,berenz2018playful,coronado2019robots}. In our work, we structure the BT to accept an action sequence assigned by the LTL planner similar to \cite{lan2019autonomous,li2021reactive}. \textcolor{black}{From an implementation perspective, our BT explicitly delegates all strategies while maintaining a simple structure that is extendable to other types of disturbances.} In addition, disturbance detection rates at different frequencies are employed by leveraging the BT's modularity property.

Our contributions are summarized as follows:
\begin{itemize}[leftmargin=*]
    \item \textcolor{black}{To handle potential disturbances from locomotion failures and environmental changes, we propose local and global task reallocation approaches given a global LTL specification. This approach eliminates the need to reconstruct the entire team model or resynthesize a new task.}
    
    
    \item \textcolor{black}{To select among reallocation strategies at run-time, we present a BT-based execution layer interfacing with low-level feedback. Our BT structure allows different types of disturbances to be detected at different rates.}
    
    


    \item We evaluate our pipeline in a simulated hospital environment with a heterogeneous robot team consisting of both \textcolor{black}{quadrupedal} and mobile robots \textcolor{black}{as shown in Fig.~\ref{fig:vision}}. \textcolor{black}{An open-source software package\footnote{\href{https://github.com/GTLIDAR/ltl\_multi\_agent}{https://github.com/GTLIDAR/ltl\_multi\_agent}.} is provided for the proposed reactive multi-robot task allocation and planning framework.}
    
\end{itemize}











\section{Preliminaries}
\label{sec:Prelim}
\subsection{LTL basics}
Linear temporal logic (LTL) has been widely used to encode temporal task specifications and automatically synthesize the system's transition behaviors. A specification $\phi$ is constructed from atomic propositions $\pi \in \Pi$, which is evaluated to be ${\rm True}$ or ${\rm False}$ and follows the syntax $\phi \: := \pi \:|\: \neg\: \phi \:|\: \phi_1 \:\wedge\: \phi_2 \:|\: \ocircle\phi \:|\: \phi_1 \:\mathcal{U}\: \phi_2 \:|\: \phi_1 \:\mathcal{R}\: \phi_2$. Boolean operators $\neg$ ``not'' and $\wedge$ ``and'' in addition to a set of temporal operators $\ocircle$ ``next'', $\mathcal{U}$ ``until'', and $\mathcal{R}$ ``release'', are denoted. To be concise, we omit the derivations of other boolean operators such as $\vee$ ``or'', $\rightarrow$ ``implies'', and $\leftrightarrow$ ``if and only if'', as well as temporal operators $\Diamond\,\phi$ ``eventually $\phi$'' and $\square\,\phi$ ``always $\phi$''. 


A common usage of LTL is for constructing an automaton. A non-deterministic automaton (NFA) is defined as a tuple $\mathcal{Q} = (S_{\mathcal{Q}}, S_{0,\mathcal{Q}}, \Sigma, \delta_{\mathcal{Q}}, F)$ such that $S_{\mathcal{Q}}$ is a set of states ($s_{\mathcal{Q}} \in S_{\mathcal{Q}}$), $S_{0,\mathcal{Q}} \subseteq S_{\mathcal{Q}}$ is a set of initial state, $\Sigma$ is the input alphabet, $\delta_{\mathcal{Q}}$ is a set of transition relations such that $\delta_{\mathcal{Q}} : S_\mathcal{Q} \times \Sigma \rightarrow 2^{S_{\mathcal{Q}}}$, and $F$ is a set of accepting final states. In addition, LTL formulas are evaluated over a sequence $\sigma: \:\mathbb{N} \:\rightarrow\: 2^\Pi$ where $\sigma(t) \:\subset\: \Pi$ represents all true propositions at time $t$ \cite{schillinger2018simultaneous}.


For this framework, a transition system (TS) is created by combining data from the topological map and the robots' operating states. A TS is defined as a tuple $\mathcal{T} = (S_{\mathcal{T}}, s_{0,\mathcal{T}}, A_{\mathcal{T}}, \Pi_\mathcal{T}, \mathcal{L})$ such that $S_{\mathcal{T}}$ is a set of system states ($s_{\mathcal{T}} \in \mathcal{S}_{\mathcal{T}}$), $s_{0,\mathcal{T}} \in \mathcal{S}_{\mathcal{T}}$ is the initial system state, $A_{\mathcal{T}}$ is a set of available system actions, $\Pi_{\mathcal{T}}$ is the set of system propositions, and $\mathcal{L}$ : $S_{\mathcal{T}} \:\rightarrow\: 2^{\Pi_{\mathcal{T}}}$ is a labeling function that assigns atomic propositions to states\cite{banks2020multi}. We use $\texttt{Succ}(s_{\mathcal{T}})=\{s_{\mathcal{T}}^{\prime} \in S_{\mathcal{T}}|(s_{\mathcal{T}},s_{\mathcal{T}}^{\prime}) \in A_{\mathcal{T}}\}$ to denote the successors of $s_{\mathcal{T}}$ and $\texttt{Pred}(s_{\mathcal{T}})=\{s_{\mathcal{T}}^{*} \in S_{\mathcal{T}}|(s_{\mathcal{T}}^{*},s_{\mathcal{T}}) \in A_{\mathcal{T}}\}$ as the predecessors of $s_{\mathcal{T}}$. By combining a TS with an NFA, a product automaton (PA) $\mathcal{P}$ composed of system states and mission specifications is generated. 
It is represented by $\mathcal{P} = \mathcal{Q} \otimes \mathcal{T} = (S_{\mathcal{P}}, S_{0,\mathcal{P}}, A_{\mathcal{P}})$ such that $S_{\mathcal{P}}=S_{\mathcal{Q}} \times S_{\mathcal{T}}$ is the set of states ($s_{\mathcal{P}} \in S_{\mathcal{P}}$), $S_{0,\mathcal{P}}=S_{0,\mathcal{Q}} \times \{s_{0,\mathcal{T}}\}$ is the set of initial states, and $A_{\mathcal{P}}=\{ ((s_{\mathcal{Q}},s_{\mathcal{T}}),(s_{\mathcal{Q}}^{\prime},s_{\mathcal{T}}^{\prime})) \in S_{\mathcal{P}} \times S_{\mathcal{P}}:(s_{\mathcal{T}},s_{\mathcal{T}}^{\prime}) \in A_{\mathcal{T}} \wedge s_{\mathcal{Q}}^{\prime} \in \delta_{\mathcal{Q}}(s_{\mathcal{Q}},\mathcal{L}(s_{\mathcal{T}}))\}$.

%
\begin{figure*}
    \centering
    \includegraphics[width=7.0in]{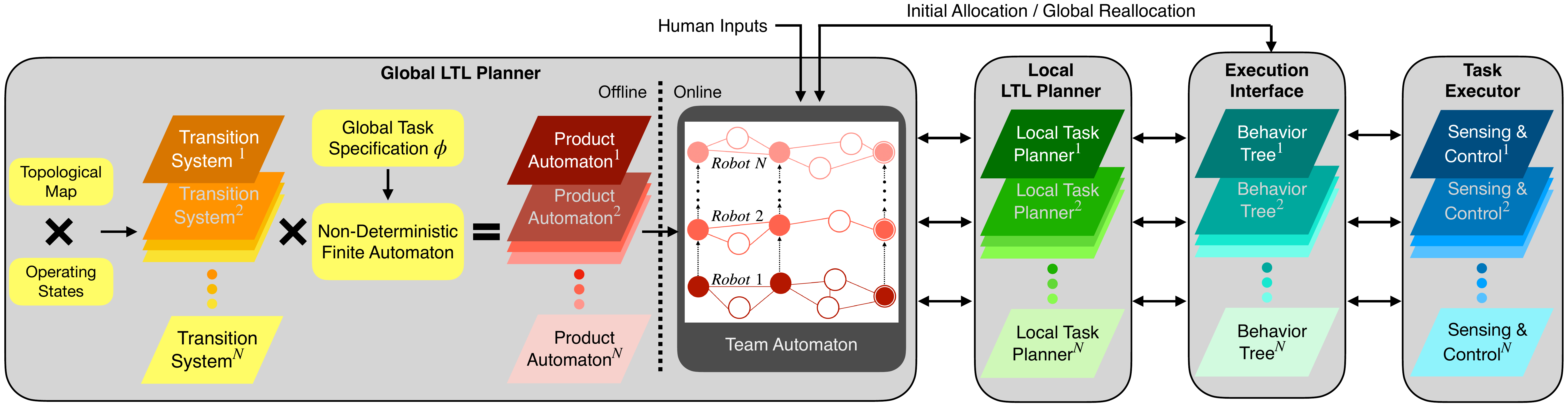}
    \caption{An overview of the hierarchical planning framework. The global LTL planner portrays the high-level LTL task planner. Once the product automaton is created, it is sent to the team automaton, where a sequence of actions are assigned to individual robots. During execution, the local task planner, the behavior tree, and the hardware on the robot will react to disturbances and handle accordingly.}
    \label{fig:system_diagram}
    \vspace{-0.2in}
\end{figure*}

\subsection{Offline task allocation}\label{subsec:offline_STAP}
We introduce the baseline task allocation method and terminologies used throughout this paper. Readers are referred to \cite{schillinger2018decomposition} for more details. Given the LTL semantics above, a global task $\phi$ along with its corresponding NFA $\mathcal{Q}$ can be specified for a whole team of $N$ agents, each of which has its own TS $\mathcal{T}^{(r)}$ and a corresponding PA $\mathcal{P}^{(r)} = \mathcal{Q} \otimes \mathcal{T}^{(r)}, r = \{1, \ldots, N\}$. Next, we will introduce a criterion that identifies the decomposed parts of $\phi$ based on the assumption that each agent executes its sub-task independently.
\begin{definition}[Finite Decomposition \cite{schillinger2018decomposition}]
Let $\mathcal{J}_r$ with $r \in \{1,\dotsc,N\}$ be a set of finite LTL task specifications and $\sigma_i$ is any sequence s.t. $\sigma_r \models \mathcal{J}_r$. These tasks are called decomposition of the global finite LTL specification $\phi$, iff:
\begin{equation}
    \sigma_{j_1}\dotsc\sigma_{j_r}\dotsc\sigma_{j_N} \models \phi
\end{equation}
for all permutations of $j_r \in \{1,\dotsc,N\}$ and all respective sequences $\sigma_r$.
\end{definition}
Based on this criterion, a decomposition set $\mathcal{D} \subseteq S_\mathcal{Q}$ can be derived from $\mathcal{Q}$, which includes all NFA states that allocate tasks. Then the action-state sequence allocated to each agent is computed by first constructing a team automaton. 
\begin{definition}[Team automaton \cite{schillinger2018decomposition}]\label{def:team_automaton}
The team automaton $\mathcal{G}$ is a union of N local PA $\mathcal{P}^{(r)}$ with $r \in \{1,\dotsc,N\}$ and defined by $\mathcal{G} := (S_{\mathcal{G}},S_{0,\mathcal{G}},F_{\mathcal{G}},A_{\mathcal{G}})$, where:
\begin{itemize}
    \item $S_{\mathcal{G}}=\{(r, s_{\mathcal{Q}}, s_{\mathcal{T}}):r \in \{1,\dotsc,N\}, (s_{\mathcal{Q}},s_{\mathcal{T}}) \in S_{\mathcal{P}}^{(r)} \}$ is the set of states;
    \item $S_{0,\mathcal{G}}=\{(r, s_{\mathcal{Q}}, s_{\mathcal{T}}):r=1, (s_{\mathcal{Q}},s_{\mathcal{T}}) \in S_{0,\mathcal{P}}^{(1)} \}$ is the set of initial states;
    \item $F_{\mathcal{G}}=\{(r, s_{\mathcal{Q}}, s_{\mathcal{T}}) \in S_{\mathcal{G}}:s_{\mathcal{Q}} \in F\}$ is the set of final accepting states;
    \item $A_{\mathcal{G}}=\bigcup_r A_{\mathcal{P}}^{(r)} \cup \zeta$ is the set of actions including switch transitions $\zeta$.
\end{itemize}
\end{definition}
We use $\delta_{\mathcal{G}}:S_{\mathcal{G}} \rightarrow S_{\mathcal{G}}$ to denote the transitions corresponding to $A_{\mathcal{G}}$. The set $\zeta \subset S_{\mathcal{G}} \times S_{\mathcal{G}}$ denotes the \textit{switch transitions}, each of which $\varsigma=((i,s_{\mathcal{Q}},s_{\mathcal{T}}),(j,s_{\mathcal{Q}}^{\prime},s_{\mathcal{T}}^{\prime}))$ is defined as a transition between two states in $\mathcal{G}$ iff: 1) $j=i+1$: connects to the next agent; 2) $s_{\mathcal{Q}}=s_{\mathcal{Q}}^{\prime}$: the NFA state remains unchanged; 3) $s_{\mathcal{T}}^{\prime}=s_{0,\mathcal{T}}^{(j)}$: points to an initial agent state; 4) $s_{\mathcal{Q}} \in \mathcal{D}$: the NFA state is inside the decomposition set.

The team automaton is a combination of every agent's PA $\mathcal{P}^{(r)}$ \footnote{For simplicity, we abuse $\mathcal{P}^{(r)}$ to denote a sub-graph in $\mathcal{G}$ which contains the same states $S_{\mathcal{P}}^{(r)}$ and transitions except, the robot index $r$ is appended.} with additional switch transitions which can reassign an agent's remaining task to another agent. \textcolor{black}{As illustrated in Fig.~\ref{fig:cartoon_reallocation}(a)}, if \textcolor{black}{a global action sequence $\beta$} on the team automaton is found, we can state that task allocation and planning have been accomplished simultaneously (namely STAP \cite{schillinger2018decomposition}). By projecting $\beta$ onto the PA of each agent, tasks can be executed in parallel. This process of finding an initial set of action-state sequence is called the \textit{offline allocation}, \textcolor{black}{whereas the state space complexity scales linearly with the number of agents. The rest of this paper will focus on addressing external disturbances during real-world deployment.}

\section{Problem Formulation}\label{problem_formulation}

\subsection{Disturbance characterization}
In this section, we categorize disturbances into four classes in order to pair a reactive strategy that can efficiently resolve the problem. Unless specified, the following disturbances apply to both legged and wheeled robots. 
\begin{itemize}[leftmargin=*]
  \item \textbf{Loss of balance} refers to a scenario where a \emph{legged robot} falls due to an unstable gait or erratic controller output.
  \item \textbf{Critical failure} refers to an irrecoverable hardware or software malfunction such as a damaged motor or a software glitch. 
  \item \textbf{Unexpected robot state change} refers to a situation where a robot detects a sudden shift in the robot's state. 
  \item \textbf{Environmental change} refers to an environmental event preventing the robot from continuing its current task.
  
\end{itemize}





\subsection{STAP reallocation}
Given the baseline STAP approach (Sec.~\ref{subsec:offline_STAP}), we seek to find an efficient reallocation strategy that is specific to each category of disturbance. For this problem, two aspects need to be investigated: 1) a formal guarantee to complete the global task; 2) a set of completed tasks by the whole team. To this end, we define a STAP reallocation problem as the following:

\begin{problem*}
Given an initial task assignment and the current TS of every agent, one finds a set of new action-state sequences for the agents to accomplish the global task without restarting the whole mission or re-synthesizing a new mission.
\end{problem*}


However, the task reallocation strategies are only compatible with high-level information such as changes in the NFA state or the TS state. Therefore, we propose a BT-based mid-level to organize different disturbance types into comprehensible inputs for the high-level planning framework. 

Fig.~\ref{fig:system_diagram} shows the planning architecture consisting of 1) a high-level task planner that performs offline allocation and online reallocation; 2) a mid-level BT interface to execute the assigned action plan for each agent; 3) low-level controllers that power the actuators on legged and wheeled robots.

\section{Reallocation Approach}\label{reallocation_approach}
\begin{figure*}
    \centering
    \includegraphics[width=\linewidth]{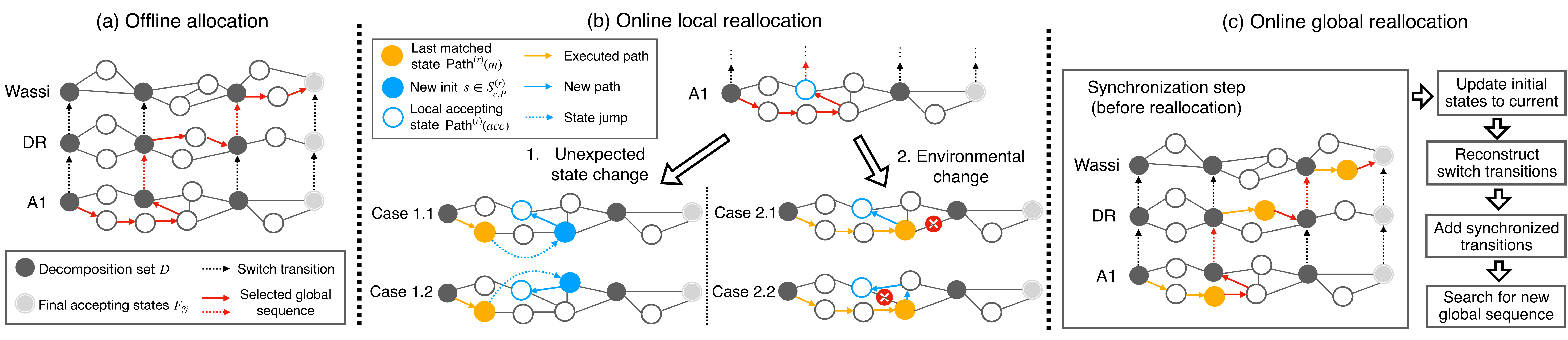}
    \caption{\textcolor{black}{Conceptual illustrations of local and global reallocations: (a) offline allocation given the team automaton. A global sequence (the red lines with arrows) is found during the offline phase as an initial task allocation. Assuming A1 undergoes a disturbance such as \textit{unexpected state change} and \textit{environmental change} as shown in subfigure (b). In correspondence to Algorithm \ref{code:local-1}, Case 1.1 refers to a scenario where the new initial state after the state jump is on the original offline-generated global sequence, while Case 1.2 corresponds to a completely new initial state and requires a replanning. Similarly in Algorithm \ref{code:local-2}, Case 2.1 demonstrates a scenario where the transition change doesn't affect the execution of the original global path, while a replanning is required in Case 2.2 due to a deleted edge on the original global path (represented by the cross marker). Subfigure (c) reveals a block diagram on the essential steps for a global reallocation. The synchronization step notifies the planner of each agent's current state (denoted by the yellow circle) and execution history to proceed with the remaining steps. More details are explained in Sec.~\ref{subsec:global_reallocation}.}}
    \label{fig:cartoon_reallocation}
    \vspace{-0.2in}
\end{figure*}

To solve the STAP reallocation problem, we propose two approaches: a local and global approach, both of which are designed at the high level. \textcolor{black}{Fig.~\ref{fig:cartoon_reallocation}(b) and \ref{fig:cartoon_reallocation}(c) show the workflow and conceptual examples of both local and global reallocation.} 

\subsection{Local reallocation addressing unexpected state changes}\label{approach:state_change}
The offline task allocation process generates an action-state sequence for each agent, which assumes every action is performed successfully. During  execution unexpected interventions could occur, which would undermine the original plan. For instance, if a human removes a load carried by the robot before the robot reaches its destination, an unforeseen robot state change occurs. 
To resolve this intervention, we introduce the local task reallocation approach. Suppose $\texttt{Path}^{(r)}$ is a planned state sequence for robot $r$, generated from the original team automaton. Let $\texttt{Path}^{(r)}({\rm acc})$ denote the agent's local accepting state and $s_{\mathcal{T}}^{\triangle}$ be the current state after the intervention. By comparing the state sequence execution history and $\texttt{Path}^{(r)}$, the last matched state is identified as $\texttt{Path}^{(r)}(m)$, whose NFA and TS are written as $\texttt{Path}^{(r)}_{\mathcal{Q}}(m)$ and $\texttt{Path}^{(r)}_{\mathcal{T}}(m)$. Thus, $\texttt{Path}^{(r)}_{\mathcal{T}}(m+1)=s_{\mathcal{T}}^{\triangle}$. Now, the problem is reformulated to find a path on the local PA, starting from an up-to-date initial set defined as 
 $S_{c,\mathcal{P}}^{(r)}=\{(r,s_{\mathcal{Q}},s_{\mathcal{T}}) \in S_{\mathcal{P}}^{(r)} | s_{\mathcal{T}}=s_{\mathcal{T}}^{\triangle}, \forall s_{\mathcal{Q}} \in \delta_{\mathcal{Q}}(\texttt{Path}_{\mathcal{Q}}^{(r)}(m),\mathcal{L}(\texttt{Path}^{(r)}_{\mathcal{T}}(m)))  \}$. The $\texttt{find-path}$ method throughout this work is performed by Dijkstra's algorithm. Note that the original path can be reused if the current state happens to be on the agent's original path. This local task reallocation approach is summarized in Algorithm \ref{code:local-1}. $\texttt{Path}^{(r)}(i:)$ denotes a sub-path starting from the $i^{\rm th}$ element.
 
 \begin{theorem}
In the presence of one robot experiencing unexpected state changes and all other agents not interfered (i.e., can successfully accomplish their tasks), the global specification $\phi$ will be fulfilled if a path is found by the local reallocation method in Algorithm \ref{code:local-1} on the local PA $\mathcal{P}^{(r)}$.

\end{theorem}

\begin{algorithm}
    \small
  \caption{Local reallocation: unexpected state change}
  \textbf{Input:} $\mathcal{P}^{(r)},\texttt{Path}^{(r)}$\\
  \textbf{Output:} A new path $\texttt{Path}^{(r)+}$
  \label{code:local-1}
  \begin{algorithmic}
    \STATE{$\texttt{Path}^{(r)+} \gets \texttt{empty path}$}
    \STATE{$\texttt{Path-set}^{(r)} \gets \texttt{empty set}$}
    \FOR{$s \ \textbf{in} \ S_{c,\mathcal{P}}^{(r)}$}
        \IF{$s \ \textbf{in} \ \texttt{Path}^{(r)}$}
            \STATE $i \gets \texttt{getIndex}(\texttt{Path}^{(r)}(s))$
            \STATE $\texttt{Path-set}^{(r)}.\text{append}(\texttt{Path}^{(r)}(i:))$
            \STATE \textbf{continue}
        \ELSE
            \STATE $\texttt{Path-set}^{(r)}.\text{append}(\texttt{find-path}(s, \texttt{Path}^{(r)}({\rm acc})))$
        \ENDIF
    \ENDFOR
    \STATE $\texttt{Path}^{(r)+} \gets \texttt{find-best}(\texttt{Path-set}^{(r)})$
  \end{algorithmic}
\end{algorithm}

\begin{proof}
\textcolor{blue}{According to Algorithm \ref{code:local-1} and the definition of $S_{c,\mathcal{P}}^{(r)}$, the new state sequence $\texttt{Path}^{(r)+}$ (i) connects to agent $r$'s executed state sequence through a valid NFA transition since $s_{\mathcal{Q}} \in \delta_{\mathcal{Q}}(\texttt{Path}_{\mathcal{Q}}^{(r)}(m),\mathcal{L}(\texttt{Path}^{(r)}_{\mathcal{T}}(m))$; and (ii) ends with the original local accepting state $\texttt{Path}^{(r)}({\rm acc})$. In other words, the originally assigned sub-task for agent $r$ is fulfilled again. Given the assumption that the rest of the agents are not interfered by any disturbances, agent $r$'s newly concatenated sequence, along with the other agents' planned sequences, consist a global action sequence $\beta$ on the team automaton $\mathcal{G}$ again. Then according to the correctness property in \cite{schillinger2018simultaneous}, the projected global path onto the NFA satisfies the mission specification $\phi$.}    
\end{proof}

\vspace{-0.1in}
\subsection{Local reallocation addressing environmental changes}
In the previous section, the disturbance shifts the robot's state but does not modify the environment, which will be addressed in this section. Such a disturbance will directly impact the TS. For instance, if the floor is occupied by an impassable object, the mobile robot would encounter a navigation failure and would not be able to transition to its next expected state. In this case, the robot will receive the changes to be made on TS called $Info(t)^{(r)}$. Each update contains three types of information: 1) $(s_{\mathcal{T}}, s_{\mathcal{T}}^{\prime}) \in Add(t)$ if $s_{\mathcal{T}}$ is allowed to transit to $s_{\mathcal{T}}^{\prime}$; 2) $(s_{\mathcal{T}}, s_{\mathcal{T}}^{\prime}) \in Delete(t)$ if $s_{\mathcal{T}}$ is not allowed to transit to $s_{\mathcal{T}}^{\prime}$; 3) $(b, s_{\mathcal{T}}) \in Relabel(t)$ if the labeling function of state $s_{\mathcal{T}}$ is updated to $b \subseteq 2^{AP}$.

The TS change can be reflected by directly modifying the team automaton using the PA revision strategy in \cite{guo2013revising}. When a single agent $r$ receives an update, the latest team automaton is revised by only updating corresponding $\mathcal{P}^{(r)}(t)$\footnote{We use $\cdot (t)$ to denote an updated automaton at time $t$ given the transition relation is changed.}. All deleted transitions, i.e., edges, are added into a set $R(t)$.

\begin{definition}[Updating rules]
\textcolor{black}{
$\mathcal{G}(t)$ (more specifically, only $\mathcal{P}^{(r)}(t)$) is updated given the $Info(t)$ from agent $r$ following the rules:}
\begin{itemize}
    \item \textcolor{black}{If $(s_{\mathcal{T}}, s_{\mathcal{T}}^{\prime}) \in Add(t)$, $(r,s_{\mathcal{Q}}^n,s_{\mathcal{T}}^{\prime})$ is in $\delta_{\mathcal{G}}((r,s_{\mathcal{Q}}^m,s_{\mathcal{T}}))$ for $\forall s_{\mathcal{Q}}^n,s_{\mathcal{Q}}^m$ satisfying $s_{\mathcal{Q}}^n \in \delta_{\mathcal{Q}}(s_{\mathcal{Q}}^m,\mathcal{L}(s_{\mathcal{T}}))$;} 
    
    \item \textcolor{black}{If $(s_{\mathcal{T}}, s_{\mathcal{T}}^{\prime}) \in Delete(t)$, $(r,s_{\mathcal{Q}}^n,s_{\mathcal{T}}^{\prime})$ is deleted from $\delta_{\mathcal{G}}((r,s_{\mathcal{Q}}^m,s_{\mathcal{T}}))$ for $\forall s_{\mathcal{Q}}^n,s_{\mathcal{Q}}^m \in S_\mathcal{Q}^{(r)}$;}
    
    \item \textcolor{black}{If $(b, s_{\mathcal{T}}) \in Relabel(t)$, then $\forall s_{\mathcal{T}}^{*} \in \texttt{Pred}(s_{\mathcal{T}})$: $(r,s_{\mathcal{Q}}^n,s_{\mathcal{T}}^{*})$ is added to $\delta_{\mathcal{G}}((r,s_{\mathcal{Q}}^m,s_{\mathcal{T}}))$ for $\forall s_{\mathcal{Q}}^n \in \delta_{\mathcal{Q}}(s_{\mathcal{Q}}^m,b)$; $(r,s_{\mathcal{Q}}^n,s_{\mathcal{T}}^{*})$ is deleted from $\delta_{\mathcal{G}}((r,s_{\mathcal{Q}}^m,s_{\mathcal{T}}))$ for $\forall s_{\mathcal{Q}}^n \notin \delta_{\mathcal{Q}}(s_{\mathcal{Q}}^m,b)$}
\end{itemize}
\end{definition}
If a disturbance was detected on an agent's TS, a new type of task reallocation algorithm is necessary. Note that no unexpected robot state is assumed in this case and the last matched state is equivalent to the current state, i.e. $\texttt{Path}^{(r)}_{\mathcal{T}}(m)=s_{\mathcal{T}}^{\triangle}$. Given the revised local PA $\mathcal{P}^{(r)}(t)$, we propose a different replanning approach in Algorithm \ref{code:local-2}, compared to the one in Sec.~\ref{approach:state_change}. 

\begin{theorem}
In the presence of environmental change and all other agents that are not interfered (i.e., can successfully accomplish their tasks), the global specification $\phi$ will be fulfilled if a path is found by the local reallocation method in Algorithm \ref{code:local-2} on the revised local PA $\mathcal{P}^{(r)}(t)$.


\end{theorem}

\begin{proof}
\textcolor{blue}{According to Algorithm \ref{code:local-2}, the new state sequence $\texttt{Path}^{(r)+}$ starts from the last state $\texttt{Path}^{(r)}(m)$ in agent $r$'s execution history and reaches the same local accepting state $\texttt{Path}^{(r)}({\rm acc})$. Since the PA updating rules preserve valid NFA transitions, the originally assigned sub-task for agent $r$ is still fulfilled by the newly found path on $\mathcal{P}^{(r)}(t)$. Same as Proposition 1, a global action sequence $\beta$ is formed assuming the rest agents are not interrupted by any disturbances. Consequently, the global specification $\phi$ is satisfied again.}
\end{proof}

\begin{algorithm}[h]
\small
  \caption{Local reallocation: environmental change}
  \textbf{Input:} $\mathcal{P}^{(r)}(t),\texttt{Path}^{(r)},Info(t)$\\
  \textbf{Output:} A new path $\texttt{Path}^{(r)+}$
  \label{code:local-2}
  \begin{algorithmic}
    \STATE{$\texttt{Path}^{(r)+} \gets \texttt{empty path}$}
    \STATE $\mathcal{P}^{(r)}(t),R(t) \gets \texttt{UpdatePA}(\mathcal{P}^{(r)}(t),Info(t))$
    \IF{$R(t) \cap \texttt{edge}(\texttt{Path}^{(r)}) \neq \varnothing$}
    \STATE $\texttt{Path}^{(r)+} \gets \texttt{find-path}(\texttt{Path}^{(r)}(m),\texttt{Path}^{(r)}({\rm acc}))$
    \ELSE
    \STATE{$\texttt{Path}^{(r)+} \gets \texttt{Path}^{(r)}(m:)$}
    \ENDIF
  \end{algorithmic}
\end{algorithm}

\vspace{-0.1in}
\subsection{Global reallocation}\label{subsec:global_reallocation}
The two aforementioned local task reallocation approaches do not consider replanning for the whole team, which results in a sub-optimal strategy. Furthermore, if the local task reallocation fails to find a new plan for the agent, a succeeding global task reallocation over the entire team is activated. First, a synchronization step will be executed where the task planner requests for each agent's current TS state and sets it to be the latest initial TS state $s_{0,\mathcal{T}}^{(r)}(t)$. Then the initial PA set $S_{0,\mathcal{P}}^{(r)}(t)$ is updated accordingly by keeping $S_{0,\mathcal{Q}}^{(r)}$ the same. Since each $\mathcal{P}^{(r)}(t)$ has been updated during local reallocation if needed, the team automaton is only modified by updating the initial set of states and switch transitions, in addition to appending the \textit{synchronized transitions}.

\begin{definition}[Synchronized team automaton]
The synchronized team model $\mathcal{R} := \mathcal{G}(t)$ is a union of $N$ product automata $\mathcal{P}^{(r)}(t)$ given the updated product states after synchronization, where $r \in \{1,\dotsc,N\}$ and $\mathcal{R}:=(S_{\mathcal{R}},S_{0,\mathcal{R}},F_{\mathcal{R}},A_{\mathcal{R}})$ consists of:
\begin{itemize}
    \item $S_{\mathcal{R}}=\{(r, s_{\mathcal{Q}}, s_{\mathcal{T}}):r \in \{1,\dotsc,N\}, (s_{\mathcal{Q}},s_{\mathcal{T}}) \in S_{\mathcal{P}}^{(r)}(t) \}$ is the set of states;
    \item $S_{0,\mathcal{R}}=\{(r, s_{\mathcal{Q}}, s_{\mathcal{T}}):r=1, (s_{\mathcal{Q}},s_{\mathcal{T}}) \in S_{0,\mathcal{P}}^{(1)}(t) \}$ is the set of initial states;
    \item $F_{\mathcal{R}}=\{(r, s_{\mathcal{Q}}, s_{\mathcal{T}}) \in S_{\mathcal{R}}:q \in F\}$ is the set of final accepting states, which remains unchanged since the NFA accepting states are fixed;
    \item $A_{\mathcal{R}}=\bigcup_r A_{\mathcal{P}}^{(r)}(t) \cup \zeta(t) \cup \xi(t)$  is the set of actions that include the updated switch transitions $\zeta(t)$ and the newly proposed synchronized transitions $\xi(t)$.
\end{itemize}
\end{definition}
Suppose $\texttt{ExePath}^{(r)}$ is the executed state sequence acquired from each agent $r$ \textcolor{black}{and $\texttt{ExePath}_{\mathcal{Q}}^{(r)}$ is the projected NFA state sequence}. The definition of a synchronized transition is as follows:  
\begin{definition}[Synchronized transition]\label{def:synchronized_transition}
The set $\xi \subset S_{\mathcal{R}} \times S_{\mathcal{R}}$ denotes synchronized transitions. Each element $\varepsilon=((i,s_{\mathcal{Q}},s_{\mathcal{T}}),(j,s_{\mathcal{Q}}^{\prime},s_{\mathcal{T}}^{\prime}))$ satisfies:
\begin{itemize}
    \item $i=j$: connects the same agent;
    \item $s_{\mathcal{Q}}=\texttt{ExePath}_{\mathcal{Q}}^{(r)}({\rm init}),s_{\mathcal{Q}}^{\prime}=\texttt{ExePath}_{\mathcal{Q}}^{(r)}({\rm final}),$ $r \in \{1,\dotsc,N\}$ starts from the initial NFA state and points to the most recent NFA nodes upon request for synchronization of each agent;
    \item $s_{\mathcal{T}}=s_{\mathcal{T}}^{\prime}$: TS state is preserved.
\end{itemize}
\end{definition}
This synchronized transition allows a new transition between two
NFA states inside each agent’s $\mathcal{P}^{(r)}(t)$. Once this is complete, each agent will be aware of the task completion status of the whole team \textcolor{black}{and avoid performing redundant tasks}. 
\textcolor{black}{
In the original team automaton \cite{schillinger2018simultaneous}, the four properties including correctness, independence, completeness, and ordered sequence are proposed to justify the rationale of finding a global path on the team automaton for a task allocation. Here we claim that our synchronized team automaton preserves these properties, so that a new global action sequence $\beta$ can be found by applying the same search algorithm performed during the offline phase (as presented in Sec.~\ref{sec:Prelim}). This process leads to a global task reallocation that assigns new sub-tasks to all agents.}



\begin{theorem}
The synchronized team automaton preserves the properties of the original team automaton, i.e., correctness, independence, completeness and ordered sequence.
\end{theorem}

\begin{proof}
\textcolor{black}{The synchronized team automaton is distinguished from the original team automaton in four folds: 1) the set of initial states $S_{0,\mathcal{R}}$ is different since the initial PA set $S_{0,\mathcal{P}}^{(1)}(t)$ is updated; 2) the set of actions $A_{\mathcal{P}}^{(r)}(t)$ from each agent $r$ is updated according to the latest $\mathcal{P}^{(r)}(t)$; 3) the switch transitions $\zeta(t)$ are removed and then reconstructed after setting the current TS state as the latest initial state $s_{0,\mathcal{T}}^{(r)}(t)$ for each agent $r$; 4) the synchronized transitions are added according to the execution history. The first three changes won't affect the properties of the team automaton in that every state $s_{\mathcal{R}} \in S_{\mathcal{R}}$ still has an NFA component $s_{\mathcal{Q}}$ constructed from $\phi$, and the switch transitions $\zeta(t)$ are reconstructed under the same definition. As for the fourth change, according to the second condition of Definition \ref{def:synchronized_transition}, new transitions between NFA states are added by connecting the initial and final NFA states from each agent $r$'s execution history $\texttt{ExePath}_{\mathcal{Q}}^{(r)}$. Although these transitions are not directly provided by $\phi$, since the states in $\texttt{ExePath}^{(r)}$ have already been traversed by agent $r$, there always exists a sequence of valid NFA transitions between $\texttt{ExePath}_{\mathcal{Q}}^{(r)}({\rm init})$ and $\texttt{ExePath}_{\mathcal{Q}}^{(r)}({\rm final})$. Therefore, synchronized transitions do not change the validness of the original NFA transitions, but only skip the executed transitions. Consequently, all of the original properties, including correctness, independence, completeness, and ordered sequence, are preserved.}      
\end{proof}

\section{BT Execution Layer}\label{bt_interface}
As described in Sec.~\ref{reallocation_approach}, both local and global task reallocation approaches enable robot recovery from various disturbances and failures. We leverage the reactivity property of BT at the middle level to rapidly select appropriate reactive strategies. 


\begin{figure}[t]
    \centering
    \includegraphics[width=3.45in]{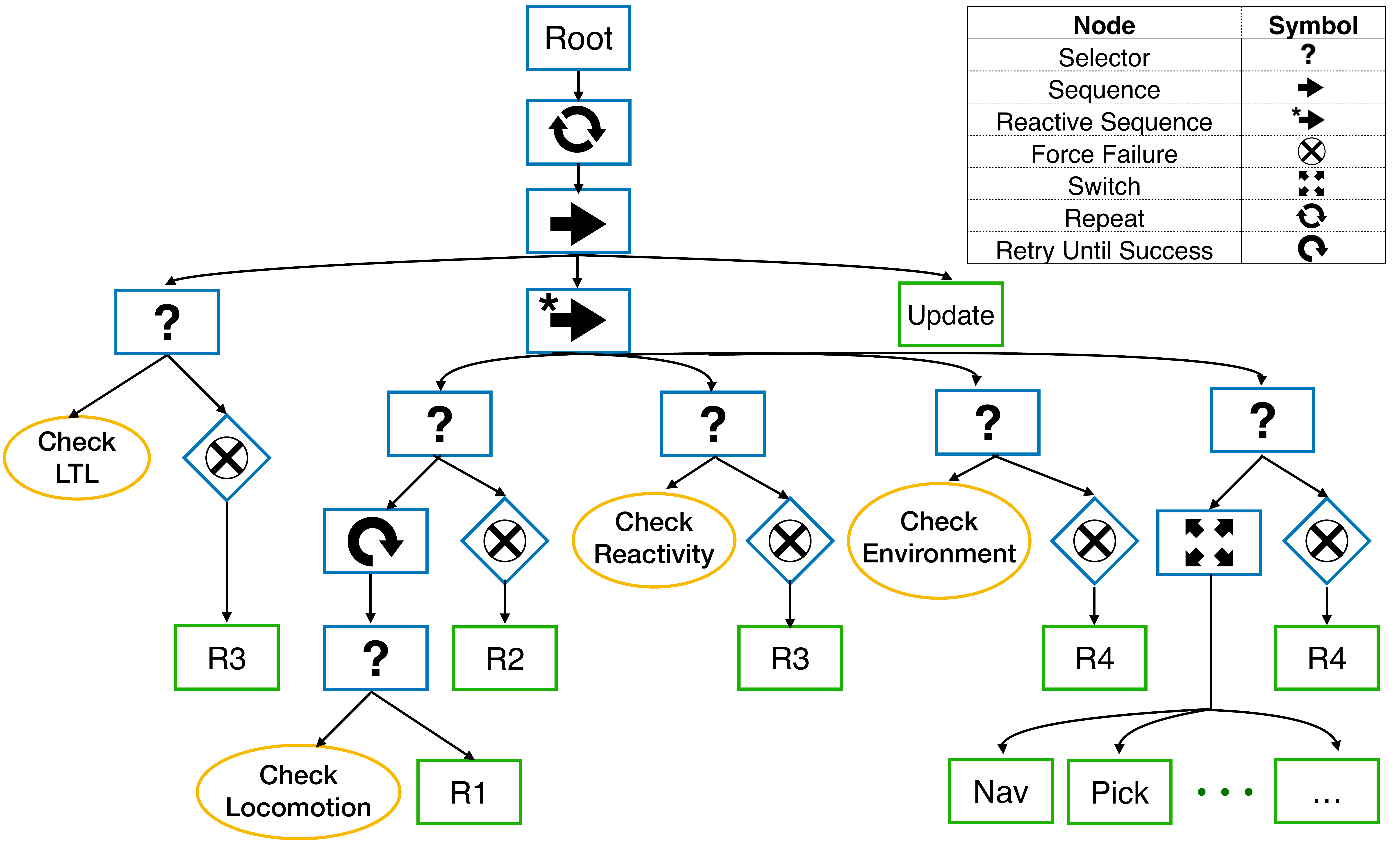}
    \caption{A behavior tree for three levels of reactive strategies is shown. Basic types of BT nodes are thoroughly described in \cite{DBLP:journals/corr/abs-1709-00084}. Yellow oval nodes are condition nodes, and green rectangular nodes are action nodes. R1 is the recovery stand for legged robots, R2 is the reallocation for critical failure, R3 is the reallocation for unexpected robot state change, and R4 is the reallocation for environmental change. Force Failure custom decorator node terminates the BT to execute an appropriate reactive strategy for the specific condition node.}
    \label{fig:bt}
    \vspace{-0.2in}
\end{figure}

\subsection{Reactive strategies}
The decision to execute one reactive strategy over another is determined by the categoried disturbances in Sec.~\ref{problem_formulation}. Four corresponding types of reactive strategies are specified below.

\begin{itemize}[leftmargin=*]
    \item \textbf{Recovery stand}. This scenario responds to a loss of balance which is specific to legged robots. In this case, the low-level controller on the robot attempts to recover from the fallen state without triggering the high-level LTL planner.
    \item \textbf{Task reallocation: critical failure}. Under this scenario, the robot is deemed incapable of working and therefore, a local task reallocation will be impractical. To resolve this issue, the robot will directly request a global task reallocation assuming a disability to transit to any TS state, to allow the remaining agents to take over its task.
    \item \textbf{Task reallocation: unexpected robot state change}. If this type of disturbance is detected, the planner will locally reallocate its task sequence without the assistance of other robots. If no local plans are feasible, the planner will perform a global task reallocation.
    \item \textbf{Task reallocation: environmental change}. If a robot encounters an environmental change, it will perform a local task reallocation. However, if a solution is not found, a global task reallocation will be performed.
    
\end{itemize}
\subsection{Behavior tree structure}
The BT structure is constructed in the form of precondition $\rightarrow$ action $\rightarrow$ effect, which is similar to the works of\cite{lan2019autonomous,li2021reactive}. This structure allows us to encode all of the reactive strategies into separate nodes within a single BT. \textcolor{black}{Two special types of nodes, Repeat and Reactive Sequence, are utilized in our BT. The former node allows a simpler BT construction instead of concatenating all actions into one big tree, while the latter node enables condition nodes to check for disturbances at varying frequencies (i.e., detection rates).} For example, as shown in Fig.~\ref{fig:bt}, checking locomotion failure, reactive state change, and environmental change would be executed continuously under the Reactive Sequence node, while checking LTL precondition would only occur prior to receiving a new task. The BT for a wheeled robot is similar but omitted due to space limits.




\section{Evaluation and Discussion}\label{evaluation}
\subsection{Experiment set-up for simulation and hardware}

To evaluate the feasibility and robustness of the proposed multi-robot task allocation and planning framework, we first establish a simulation of a hospital environment in Gazebo\cite{IgnitionFuel-OpenRobotics-Hospital} and create a topological map for defining the TS, as shown in Fig.~\ref{fig:map_hospital}. The simulation architecture is composed of a high-level LTL planning layer based on a ROS package from \cite{baran2021rosltl}, a mid-level execution interface using BehaviorTree.CPP\cite{BehaviorTreeCPP}, and a low-level navigation and controller layer using ROS navigation stack and appropriate controllers for each robot model. A convex model predictive controller from MIT Mini Cheetah \cite{di2018dynamic,kim2019highly} is used to control a Unitree A1 quadruped over rough terrain, while a conventional holonomic drive model is used on the UBTECH DR and Wassi robot. For global path planning, the robot navigates between regions using the A* algorithm \cite{hart1968formal} and performs collision avoidance with the dynamic window approach \cite{fox1997dynamic}.

\begin{figure}[t]
    \centering
    \includegraphics[width=3.5in]{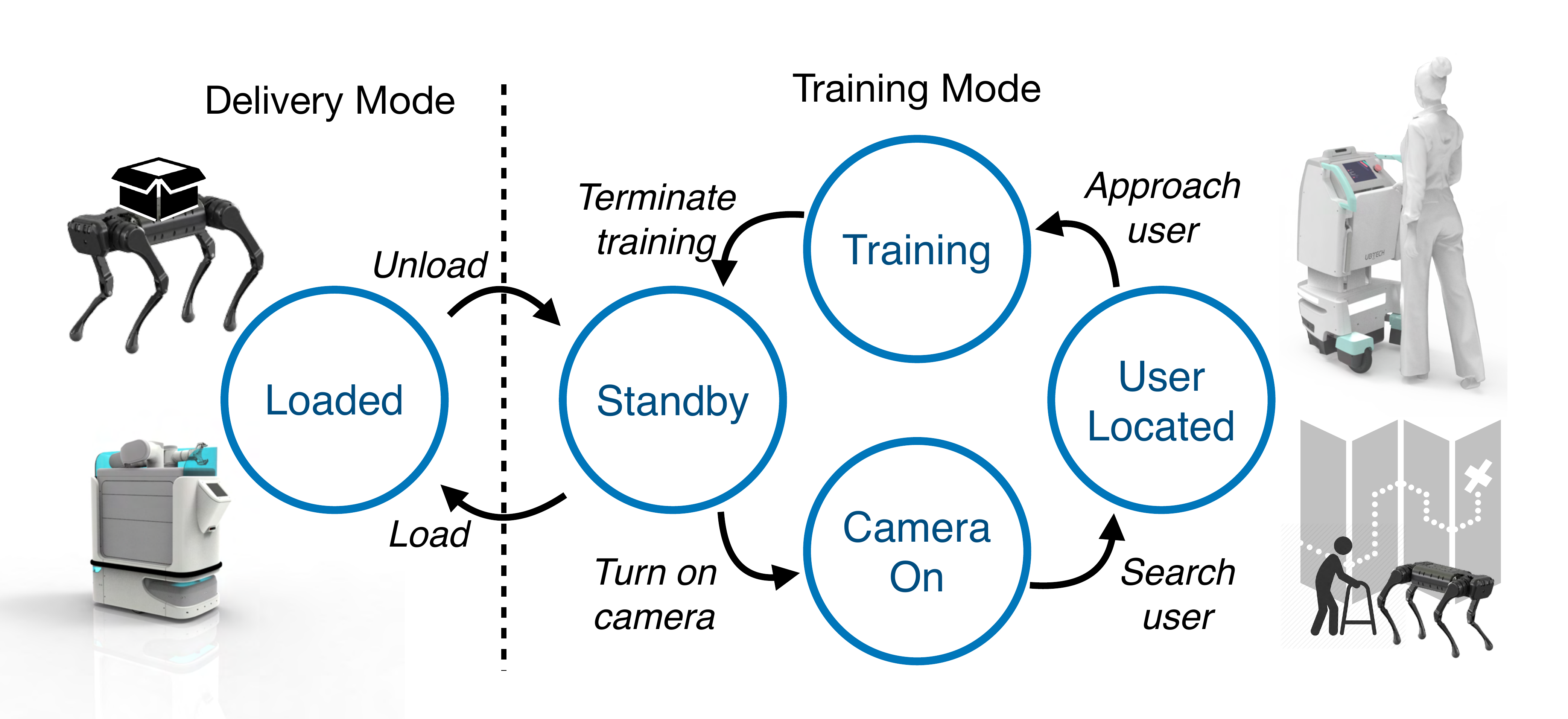}
    \caption{Operating state diagram for legged robot capable of both delivery and training task. Each node refers to robot operating states, whereas each edge refer to an action performed by the robot. Unitree A1 quadruped \emph{(top left)} and UBTECH DR \emph{(bottom left)} are shown as delivery robots, and A1 and UBTECH Wassi \emph{(top right)} are shown as training robots.}
    \label{fig:robot_definition}
    \vspace{-0.2in}
\end{figure}

\subsection{Case study}

We evaluate our framework on a heterogeneous team of robots consisting of a delivery robot DR, a walk training robot Wassi, and a quadrupedal robot A1 with both capabilities. As shown in Fig.~\ref{fig:robot_definition}, a delivery robot consists of two simple operating states (\emph{Loaded, Standby}), where the \emph{Standby} state is equivalent to \emph{Unloaded State}. Likewise, a training robot consists of 4 operating states (\emph{Standby, Camera On, User Located, Training}), where the robot visually locates and helps seniors who need walking training assistance. These operating states are encoded into TS for each type of robot.

We conduct a series of case studies in a hospital environment simulation to evaluate the reactive strategies proposed in Sec.~\ref{bt_interface}. The global mission is defined as:
\begin{figure}[t]
    \centering
    \includegraphics[width=\linewidth]{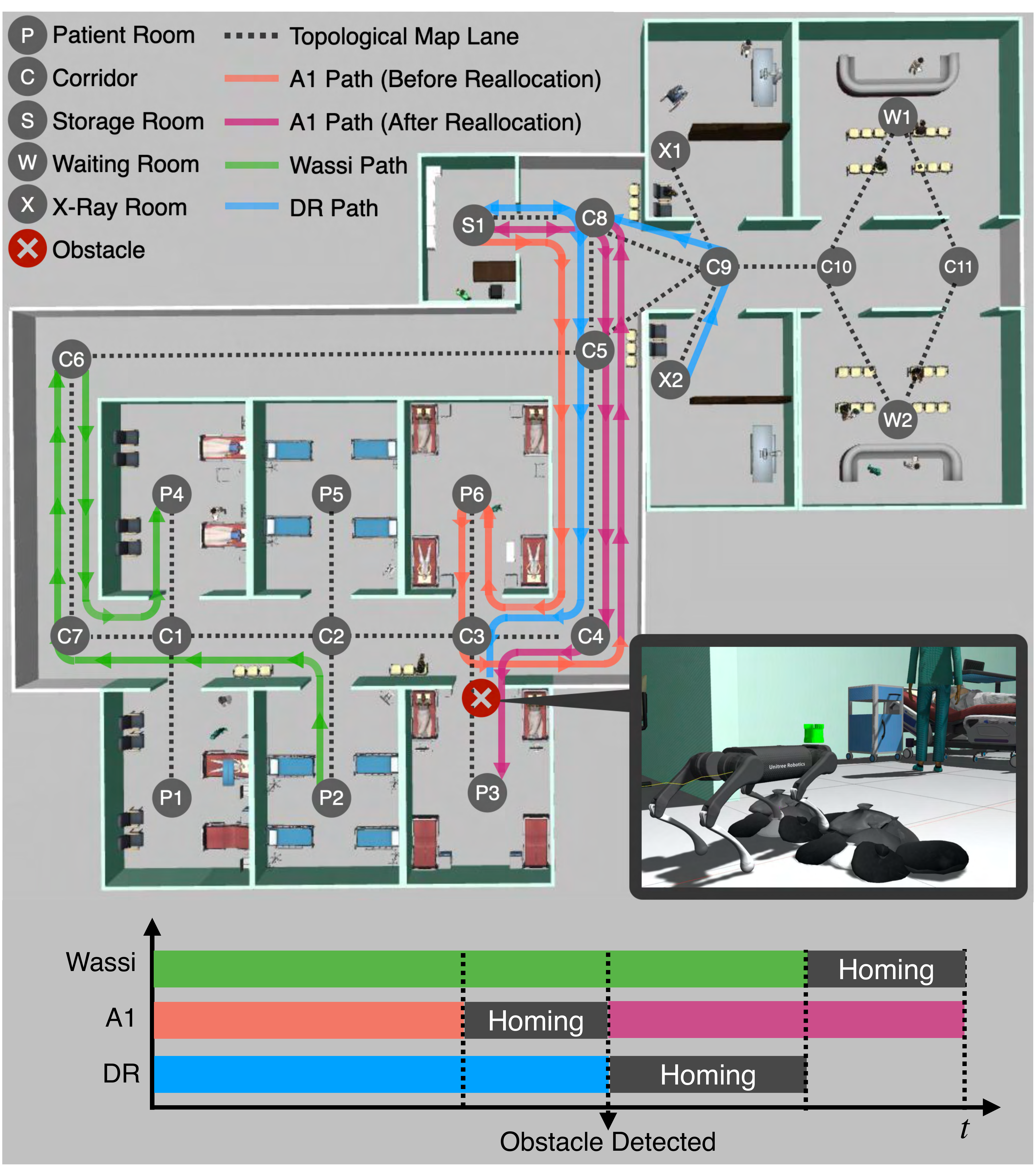}
    \caption{Topological map of a hospital environment consisting of locations that are explicitly defined in each robot's transition system. The execution history for each robot is depicted for achieving global task $\phi$ under a scenario where an unexpected garbage heap is detected. The timeline for task execution and reallocation is illustrated at the bottom.}
    \label{fig:map_hospital}
    \vspace{-0.15in}
\end{figure}

\begin{scenario}
``Deliver medicines to locations p3 and p6; meet a patient at c1; complete a walk training along the corridor between c1 and c6, and then send the patient back to p4."
\begin{subequations}
\begin{align*}
    \phi_1 = & \Diamond(p3 \wedge Standby) \wedge \square\,((\neg\,p3 \wedge \ocircle \,p3) \rightarrow Loaded) \\
    & \wedge \Diamond(p6 \wedge Standby) \wedge \square\,((\neg\,p6 \wedge \ocircle \,p6) \rightarrow Loaded) \\
    & \wedge \Diamond(p4 \wedge Standby) \wedge \square\,((\neg\,p4 \wedge \ocircle \,p4) \rightarrow Training)\\
    & \wedge \Diamond(c1 \wedge Training \wedge \Diamond\,(c7 \wedge Training \wedge \Diamond\,(c6 \wedge Tr \\
    &\text{-}aining \wedge \Diamond\,(c7 \wedge Training \wedge \Diamond\,(c1 \wedge Training)))))
\end{align*}
\end{subequations}
\end{scenario}

\begin{table}[t]
\caption{\centering Triggered times and averaged computation time for four different
recovery strategies R1 -- R4. }
\label{table_10_trials}
\begin{center}
\begin{tabular}{|c||c||c||c||c|}
    \hline
     & R1 & R2 & R3 & R4\\
    \hline
    Triggered times & 18 & 13 & 10 & 16\\
    \hline
    Local reallocation time (s) & - & - & 0.023 & 0.018\\
    \hline
    Global reallocation time (s) & - & 3.31 & 2.93 & 3.42\\
    
    \hline
\end{tabular}
\end{center}
\vspace{-0.15in}
\end{table}

Three robots, A1, DR and Wassi, are placed at various locations in the hospital before the start of the simulation. Next, the task planner decomposes the specification and assigns sub-tasks to each robot offline, \textcolor{black}{which takes 21 seconds in total. 92\% of the computation time is taken by the generation of PA for each robot, which only needs to be performed once.}

During the simulation, each robot will encounter a disturbance. Since the integration with exteroceptive systems is out of this paper's scope, all the disturbances except the locomotion failure are triggered by manually setting the checkers in BT to be false. A diagram of this simulation is displayed in Fig.~\ref{fig:map_hospital}. More details can be found in the video\footnote{\href{https://youtu.be/AO7Ms3GKzYQ}{https://youtu.be/AO7Ms3GKzYQ}}.    

\emph{1)} External force is applied to A1 and induces a loss of balance. A handcrafted whole-body recovery stand trajectory is tracked by a PD controller to assist A1 to resume its task.   

\emph{2)} A critical failure is induced for Wassi when performing the walk training task from $c7$ to $c6$. A global reallocation strategy assigns A1 to complete its delivery task first, and then proceeds to finish Wassi's incomplete walk training task.

\emph{3)} The \emph{Loaded} state of DR suddenly becomes a \emph{Standby} state. This simulates a situation in which the robot unexpectedly loses its cargo. A local reallocation succeeds in instructing the robot to return to $s1$ and pick up another cargo. Then DR is instructed to complete the original delivery task. 

\emph{4)} A garbage heap is placed in front of $p3$ to simulate an environmental change. This obstruction can only be traversed by the legged robot A1. The routes taken by each robot and the timeline for obstacle detection and task allocation are portrayed in Fig.~\ref{fig:map_hospital}. While performing its task, DR encounters the obstruction and fails to find an alternate plan via local reallocation. While A1 is returning home after completing the delivery task to $p6$, it is assigned to take over DR's incomplete delivery task at $c4$. As a result, A1 goes to $s1$ to retrieve the object for delivery, and completes the task by traveling to $p3$.

\textcolor{black}{In addition to applying individual disturbances, we also evaluate the the same scenario ten times with multiple manually triggered adversarial disturbances. Furthermore, the initial configuration for each robot is modified to prompt a different offline allocation result, for evaluating the generalization of our approach. At run-time, signals indicating robot state and environmental changes are sent to each robot's BT and prompt their reactive behaviors. 
As shown in Table~\ref{table_10_trials}, we report the number of times that the reactivity strategy is triggered and the average computation time it takes for each trial. The LTL planner is not responsible for locomotion failure (R1) and the critical failure (R2) can only be handled at global level. 
Although a global task reallocation in principle could result in an optimal task sequence according to Proposition 3, it is observed to be more computational expensive compared with the local reallocation. This is caused by a communication among all agents and extra computation for constructing the synchronized team automaton. Our framework demonstrates an 80\% success rate of completing the global mission $\phi$. The 20\% failed cases are caused by mission-level failures when more than one robot undergoes a critical failure, which causes the remaining task to be  unachievable. 
}


\begin{figure}[h]
    \centering
    \includegraphics[width=0.6\linewidth]{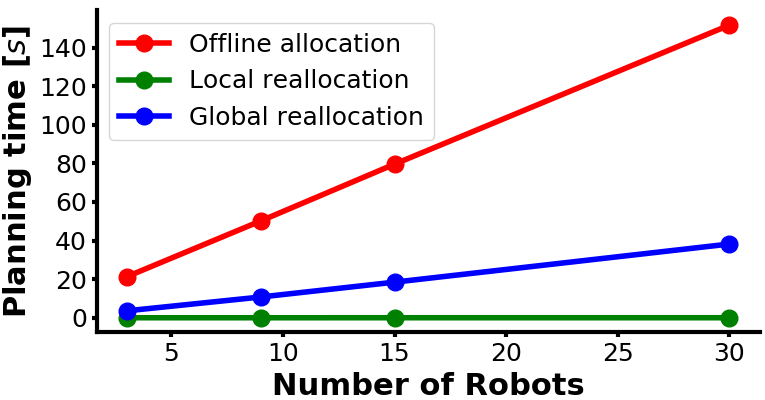}
    \caption{The recorded planning time for offline allocation, local and global reallocation given different numbers of robots.}
    \label{fig:scalability}
\end{figure}

\textcolor{black}{To evaluate the scalability of our proposed framework, we deploy 3 (described above), 9, 15, and 30 robots in the same aforementioned scenario. Due to the high computation demand from the quadruped controller, we don't run the simulation but directly trigger the LTL planner to perform reallocation strategies. In all scalability tests, the delivery robot is assumed to encounter an abrupt state change that triggers local reallocation and a critical failure that triggers global reallocation. Without significant code optimization, Fig.~\ref{fig:scalability} reveals a linear complexity for both offline allocation and global reallocation, while the time required for local reallocation maintains at the same level. An alternative but \textit{ad hoc} way for reallocation is to reconstruct the entire team model with an updated specification, which is similar to repeating the expensive offline process and requires human designer knowledge in the loop.
}

\section{Conclusion and Discussions}
In this work, we present a heterogeneous, multi-robot task allocation and planning framework equipped with a hierarchically reactive mechanism from extensive disturbances. A local and global task reallocation is performed at the high level where an LTL-based team automaton is generated to follow a formal guarantee. At the middle level, a BT framework is incorporated to promptly select different replanning strategies which can be executed at different rates. Lastly, all the work mentioned is showcased in a dynamic simulation of a hospital scenario involving quadrupeds and wheeled robots. 

\textcolor{black}{In certain scenarios, a global task reallocation could result in a more optimal task sequence than the one generated from a local reallocation. For instance, if a robot encounters a blocked path, a local strategy will first attempt to find a detour. However, if the task could be transferred over to a different robot closer to the destination, the global task reallocation will outperform the local one. 
Our current framework does not consider optimality over computational efficiency. For our future work, we will incorporate a method to take into account optimality to determine whether replanning for the whole team over a single robot is favourable.}


\bibliographystyle{IEEEtran}
\bibliography{references}

\end{document}